\documentclass[runningheads,a4paper]{llncs}
\usepackage[utf8]{inputenc}
\usepackage{helvet}
\usepackage{amssymb}
\usepackage{graphicx}
\usepackage{tikz}
\usepackage{hyperref}
\usepackage[T1]{fontenc}
\usepackage{caption}
\usepackage{subcaption}
\usepackage{siunitx}
\usepackage{cite}
\usepackage[section]{placeins}

\usepackage{url}
\urldef{\mailsa}\path|alexandre.de.brebisson@umontreal.ca|
\urldef{\mailsb}\path|alex.auvolat@ens.fr|
\urldef{\mailsc}\path|esimon@esimon.eu|
\urldef{\mailsd}\path|vincentp@iro.umontreal.ca|

\newcommand{\taxiglyph}{
	\begin{tikzpicture}[y=0.80pt, x=0.80pt, yscale=-0.030000, xscale=0.030000, inner sep=0pt, outer sep=0pt,
		draw=black,fill=black,line join=miter,line cap=rect,miter limit=10.00,line width=0.800pt]
		\path[fill=black] (284.0625,216.1406) .. controls (284.0625,217.7343) and
		(283.9219,219.9843) .. (283.6406,222.8906) .. controls (283.2656,226.0781) and
		(283.0781,228.3281) .. (283.0781,229.6406) .. controls (282.6094,243.3281) and
		(281.2031,253.6406) .. (278.8594,260.5781) .. controls (275.8594,263.3906) and
		(272.3906,265.0313) .. (268.4531,265.5000) .. controls (269.2031,265.4063) and
		(265.3125,265.3594) .. (256.7812,265.3594) .. controls (248.7187,280.4531) and
		(237.1875,288.0000) .. (222.1875,288.0000) .. controls (208.0312,288.0000) and
		(196.8281,280.4063) .. (188.5781,265.2188) -- (104.7656,265.2188) .. controls
		(97.0781,280.4063) and (85.7344,288.0000) .. (70.7344,288.0000) .. controls
		(54.7969,288.0000) and (43.5937,280.4063) .. (37.1250,265.2188) --
		(19.9688,265.2188) .. controls (19.0313,264.4688) and (17.7656,263.2969) ..
		(16.1719,261.7031) .. controls (13.0781,257.7656) and (11.5312,249.2813) ..
		(11.5312,236.2500) .. controls (11.5312,209.4375) and (21.5156,190.8281) ..
		(41.4844,180.4219) .. controls (43.8281,179.2032) and (46.1250,178.5938) ..
		(48.3750,178.5938) -- (83.5312,178.5938) .. controls (91.5937,169.9687) and
		(103.9219,157.2656) .. (120.5156,140.4844) .. controls (122.6719,138.4219) and
		(124.9219,137.0156) .. (127.2656,136.2656) .. controls (130.5469,135.1406) and
		(138.0937,134.4375) .. (149.9062,134.1562) -- (150.6094,133.4531) .. controls
		(150.6094,133.0781) and (151.6406,123.4687) .. (153.7031,104.6250) --
		(173.9531,104.6250) -- (176.7656,134.4375) -- (199.6875,134.4375) .. controls
		(202.8750,134.4375) and (207.7032,135.3750) .. (214.1719,137.2500) .. controls
		(221.7656,139.4063) and (228.1406,143.3906) .. (233.2969,149.2031) .. controls
		(235.6406,151.9218) and (238.8750,156.8437) .. (243.0000,163.9688) .. controls
		(246.0937,169.1250) and (248.8125,172.0781) .. (251.1562,172.8281) .. controls
		(252.7500,173.2968) and (256.9688,173.5312) .. (263.8125,173.5312) .. controls
		(268.0312,173.5312) and (271.0312,173.8125) .. (272.8125,174.3750) .. controls
		(274.0312,174.7500) and (275.4609,175.8281) .. (277.1016,177.6094) .. controls
		(278.7422,179.3906) and (279.5625,180.8906) .. (279.5625,182.1094) --
		(279.5625,202.5000) .. controls (282.5625,205.3125) and (284.0625,209.8594) ..
		(284.0625,216.1406) -- cycle(271.8281,210.5156) .. controls
		(271.8281,209.2031) and (271.7578,207.1875) .. (271.6172,204.4688) .. controls
		(271.4765,201.7500) and (271.4062,199.7344) .. (271.4062,198.4219) .. controls
		(271.4062,196.0781) and (271.5468,194.0625) .. (271.8281,192.3750) --
		(268.3125,192.3750) .. controls (267.1875,192.3750) and (265.8516,193.7578) ..
		(264.3047,196.5234) .. controls (262.7578,199.2891) and (261.9844,201.3281) ..
		(261.9844,202.6406) .. controls (261.9844,204.4219) and (263.4844,207.0469) ..
		(266.4844,210.5156) -- (271.8281,210.5156) -- cycle(169.5938,134.2969) ..
		controls (169.1250,128.9531) and (168.3750,120.8906) .. (167.3438,110.1094) --
		(159.7500,110.1094) -- (157.5000,134.2969) -- (169.5938,134.2969) --
		cycle(274.6406,236.3906) -- (274.6406,233.8594) -- (253.6875,233.8594) ..
		controls (257.1562,240.5157) and (258.7969,246.7500) .. (258.6094,252.5625) --
		(269.4375,252.5625) .. controls (272.9062,250.4062) and (274.6406,245.0156) ..
		(274.6406,236.3906) -- cycle(269.1562,184.9219) .. controls
		(268.4062,184.3594) and (267.9843,183.7500) .. (267.8906,183.0938) .. controls
		(255.9843,183.0938) and (248.6015,182.6485) .. (245.7422,181.7578) .. controls
		(242.8828,180.8672) and (240.6094,178.9688) .. (238.9219,176.0625) .. controls
		(237.4219,173.4375) and (235.9688,170.8125) .. (234.5625,168.1875) .. controls
		(224.8125,153.3750) and (213.5625,145.9688) .. (200.8125,145.9688) --
		(132.8906,145.9688) .. controls (131.1094,145.9688) and (129.2344,146.7188) ..
		(127.2656,148.2188) .. controls (125.2969,149.7188) and (117.5625,157.2657) ..
		(104.0625,170.8594) .. controls (91.3125,183.7031) and (84.7031,190.1250) ..
		(84.2344,190.1250) -- (50.6250,190.1250) .. controls (47.0625,190.1250) and
		(43.6406,192.0937) .. (40.3594,196.0312) .. controls (42.7031,196.2187) and
		(45.2813,197.2031) .. (48.0938,198.9844) -- (257.3438,198.9844) .. controls
		(257.6250,196.4531) and (259.0312,193.7109) .. (261.5625,190.7578) .. controls
		(264.0938,187.8047) and (266.6250,185.8594) .. (269.1562,184.9219) --
		cycle(244.8281,252.2812) .. controls (244.8281,246.1875) and
		(242.6250,241.0312) .. (238.2188,236.8125) .. controls (233.8125,232.5938) and
		(228.5625,230.4844) .. (222.4688,230.4844) .. controls (216.3750,230.4844) and
		(211.1015,232.5938) .. (206.6484,236.8125) .. controls (202.1953,241.0312) and
		(199.9688,246.1875) .. (199.9688,252.2812) .. controls (199.9688,258.2812) and
		(202.2657,263.5078) .. (206.8594,267.9609) .. controls (211.4531,272.4140) and
		(216.7500,274.6406) .. (222.7500,274.6406) .. controls (228.5625,274.6406) and
		(233.6953,272.3672) .. (238.1484,267.8203) .. controls (242.6015,263.2734) and
		(244.8281,258.0937) .. (244.8281,252.2812) -- cycle(275.7656,221.3438) ..
		controls (275.7656,218.9063) and (275.5312,216.9844) .. (275.0625,215.5781) ..
		controls (271.6875,215.6719) and (270.0000,215.7188) .. (270.0000,215.7188) ..
		controls (262.3125,215.7188) and (258.0938,212.2032) .. (257.3438,205.1719) --
		(49.2188,205.1719) .. controls (48.3750,210.4219) and (45.2812,215.3203) ..
		(39.9375,219.8672) .. controls (34.5938,224.4141) and (29.2031,226.9687) ..
		(23.7656,227.5312) -- (44.5781,227.5312) .. controls (52.3594,219.9375) and
		(61.0781,216.1406) .. (70.7344,216.1406) .. controls (81.8907,216.1406) and
		(90.6094,219.9844) .. (96.8906,227.6719) -- (196.0312,227.6719) .. controls
		(204.0000,219.9844) and (212.8125,216.1406) .. (222.4688,216.1406) .. controls
		(232.1250,216.1406) and (240.8906,219.9844) .. (248.7656,227.6719) --
		(275.6250,227.6719) .. controls (275.6250,226.9219) and (275.6250,225.8438) ..
		(275.6250,224.4375) .. controls (275.7187,222.9375) and (275.7656,221.9063) ..
		(275.7656,221.3438) -- cycle(191.6719,234.0000) -- (101.6719,234.0000) ..
		controls (103.8281,237.0937) and (105.7031,243.2812) .. (107.2969,252.5625) --
		(186.7500,252.5625) .. controls (186.7500,244.4062) and (188.3906,238.2187) ..
		(191.6719,234.0000) -- cycle(45.2812,204.4688) .. controls (45.2812,203.2500)
		and (45.0468,202.3594) .. (44.5781,201.7969) -- (35.7188,201.7969) .. controls
		(31.0313,201.7969) and (26.8594,208.0313) .. (23.2031,220.5000) --
		(25.1719,220.5000) .. controls (26.9532,220.5000) and (28.6407,220.2656) ..
		(30.2344,219.7969) .. controls (32.8594,219.0469) and (36.0235,216.9141) ..
		(39.7266,213.3984) .. controls (43.4297,209.8828) and (45.2812,206.9063) ..
		(45.2812,204.4688) -- cycle(92.8125,252.2812) .. controls (92.8125,246.2812)
		and (90.6328,241.1484) .. (86.2734,236.8828) .. controls (81.9141,232.6172)
		and (76.7344,230.4844) .. (70.7344,230.4844) .. controls (64.5469,230.4844)
		and (59.3203,232.5703) .. (55.0547,236.7422) .. controls (50.7890,240.9141)
		and (48.6562,246.0937) .. (48.6562,252.2812) .. controls (48.6562,258.2812)
		and (50.8359,263.5078) .. (55.1953,267.9609) .. controls (59.5547,272.4140)
		and (64.7344,274.6406) .. (70.7344,274.6406) .. controls (76.5469,274.6406)
		and (81.6797,272.3672) .. (86.1328,267.8203) .. controls (90.5859,263.2734)
		and (92.8125,258.0937) .. (92.8125,252.2812) -- cycle(39.6562,233.8594) --
		(21.5156,233.8594) -- (21.5156,245.1094) .. controls (21.5156,244.9219) and
		(22.1250,247.4062) .. (23.3438,252.5625) -- (34.7344,252.5625) .. controls
		(35.1094,244.3125) and (36.7500,238.0781) .. (39.6562,233.8594) --
		cycle(217.1250,181.9688) .. controls (217.1250,185.7188) and
		(216.1406,188.4375) .. (214.1719,190.1250) -- (167.0625,190.1250) .. controls
		(165.8438,188.8125) and (165.2344,187.2656) .. (165.2344,185.4844) .. controls
		(165.2344,187.0781) and (165.8438,177.2813) .. (167.0625,156.0938) .. controls
		(169.4062,153.5625) and (172.5000,152.2969) .. (176.3438,152.2969) --
		(186.6094,152.2969) .. controls (194.8594,152.2969) and (201.2813,152.8125) ..
		(205.8750,153.8438) .. controls (209.9063,154.7813) and (212.9063,158.7656) ..
		(214.8750,165.7969) .. controls (216.3750,170.9531) and (217.1250,176.3437) ..
		(217.1250,181.9688) -- cycle(159.1875,157.5000) .. controls
		(159.1875,160.5937) and (158.5313,170.6719) .. (157.2188,187.7344) .. controls
		(156.7500,188.3907) and (156.0469,189.1875) .. (155.1094,190.1250) --
		(102.3750,190.1250) .. controls (101.1563,189.0000) and (100.5469,187.9219) ..
		(100.5469,186.8906) .. controls (100.5469,186.1406) and (104.2031,181.9219) ..
		(111.5156,174.2344) .. controls (115.0781,170.0156) and (118.6875,165.8437) ..
		(122.3438,161.7188) .. controls (127.5938,155.8125) and (131.5782,152.5781) ..
		(134.2969,152.0156) .. controls (134.9531,151.9219) and (135.7031,151.8750) ..
		(136.5469,151.8750) .. controls (137.3906,151.8750) and (138.5859,151.9922) ..
		(140.1328,152.2266) .. controls (141.6797,152.4609) and (142.8750,152.5781) ..
		(143.7188,152.5781) .. controls (144.2813,152.5781) and (145.0781,152.5312) ..
		(146.1094,152.4375) .. controls (147.0469,152.4375) and (147.7969,152.4375) ..
		(148.3594,152.4375) .. controls (155.5781,152.4375) and (159.1875,154.1250) ..
		(159.1875,157.5000) -- cycle(171.5625,225.7031) -- (166.7812,225.7031) --
		(166.7812,207.2812) -- (171.5625,207.2812) -- (171.5625,225.7031) --
		cycle(165.0938,225.7031) -- (159.4688,225.7031) -- (155.9531,219.9375) --
		(152.2969,225.7031) -- (142.5938,225.7031) -- (141.3281,221.9062) --
		(135.7031,221.9062) -- (134.2969,225.7031) -- (129.7969,225.7031) --
		(135.9844,207.2812) -- (140.9062,207.2812) -- (147.3750,225.4219) --
		(153.0000,216.2812) -- (147.5156,207.2812) -- (152.7188,207.2812) --
		(156.2344,212.7656) -- (159.4688,207.2812) -- (164.8125,207.2812) --
		(159.0469,216.1406) -- (165.0938,225.7031) -- cycle(131.9062,210.5156) --
		(126.7031,210.5156) -- (126.7031,225.7031) -- (121.9219,225.7031) --
		(121.9219,210.6562) -- (116.4375,210.6562) -- (116.4375,207.2812) --
		(131.9062,207.2812) -- (131.9062,210.5156) -- cycle(140.0625,218.3906) --
		(138.3750,212.7656) -- (136.5469,218.3906) -- (140.0625,218.3906) -- cycle;
	\end{tikzpicture}
}

\newcommand{\taxi}{ \kern-0.7em \taxiglyph \kern-0.7em }

\makeatletter
\renewcommand*{\@fnsymbol}[1]{\ifcase#1\or {\protect\taxi} \else\@ctrerr\fi}
\makeatother

\begin{document}

\mainmatter 

\title{Artificial Neural Networks Applied to\\Taxi Destination Prediction}

\titlerunning{Artificial Neural Networks Applied to Taxi Destination Prediction}

%
%
\author{Alexandre de Brébisson\inst{1} \thanks{Random order, this does not reflect the weights of contributions.}
\and Étienne Simon\inst{2} \footnotemark[1]
\and Alex Auvolat\inst{3} \footnotemark[1]
\and \\ Pascal Vincent\inst{1}\inst{4} \and Yoshua Bengio\inst{1}\inst{4}.}

\authorrunning{Artificial Neural Networks Applied to Taxi Destination Prediction}

\institute{MILA lab, University of Montréal,\\
\mailsa, \\ \mailsd\\
\and
ENS Cachan,\\
\mailsc\\
\and
ENS Paris,\\
\mailsb\\
\and
CIFAR.}

\toctitle{Lecture Notes in Computer Science}
\tocauthor{Authors' Instructions}
\maketitle

\begin{abstract}
We describe our first-place solution to the ECML/PKDD discovery challenge on taxi destination prediction. The task consisted in predicting the destination of a taxi based on the beginning of its trajectory, represented as a variable-length sequence of GPS points, and diverse associated meta-information, such as the departure time, the driver id and client information. Contrary to most published competitor approaches, we used an almost fully automated approach based on neural networks and we ranked first out of 381 teams. The architectures we tried use multi-layer perceptrons, bidirectional recurrent neural networks and models inspired from recently introduced memory networks. Our approach could easily be adapted to other applications in which the goal is to predict a fixed-length output from a variable-length sequence.
\end{abstract}

\newpage

\section{Introduction}

The taxi destination prediction challenge was organized by the 2015\linebreak ECML/PKDD conference\footnote{\url{http://www.geolink.pt/ecmlpkdd2015-challenge/}} and proposed as a Kaggle competition\footnote{\url{https://www.kaggle.com/c/pkdd-15-predict-taxi-service-trajectory-i}}. It consisted in predicting the destinations (latitude and longitude) of taxi trips based on initial partial trajectories (which we call \emph{prefixes}) and some meta-information associated to each ride. Such prediction models could help to dispatch taxis more efficiently.

The dataset is composed of all the complete trajectories of 442 taxis running in the city of Porto (Portugal) for a complete year (from 2013-07-01 to 2014-06-30). The training dataset contains 1.7 million datapoints, each one representing a complete taxi ride and being composed of the following attributes\footnote{The exact list of attributes for each trajectory can be found here \url{https://www.kaggle.com/c/pkdd-15-predict-taxi-service-trajectory-i/data}}:

\begin{itemize}
	\item the complete taxi ride: a sequence of GPS positions (latitude and longitude) measured every 15 seconds. The last position represents the destination and different trajectories have different GPS sequence lengths.
	\item metadata associated to the taxi ride:
		\begin{itemize}
		\item if the client called the taxi by phone, then we have a client ID. If the client called the taxi at a taxi stand, then we have a taxi stand ID. Otherwise we have no client identification,
		\item the taxi ID,
		\item the time of the beginning of the ride (unix timestamp).
		\end{itemize}
\end{itemize}

In the competition setup, the testing dataset is composed of 320 partial trajectories, which were created from five snapshots taken at different timestamps. This testing dataset is actually divided in two subsets of equal size: the public and private test sets. The public set was used through the competition to compare models while the private set was only used at the end of the competition for the final leaderboard.

Our approach uses very little hand-engineering compared to those published by other competitors. It is almost fully automated and based on artificial neural networks. Section~\ref{sec_winning} introduces our winning model, which is based on a variant of a multi-layer perceptron (MLP) architecture. Section~\ref{sec_alternatives} describes more sophisticated alternative architectures that we also tried. Although they did not perform as well as our simpler winning model for this particular task, we believe that they can provide further insight on how to apply neural networks to similar tasks. Section~\ref{sec_exp} and Section~\ref{sec_dis} compares and analyses our various models quantitatively and qualitatively on both the competition testing set and a bigger custom testing set. The source code and instructions to reproduce our results can be found online (\url{https://github.com/adbrebs/taxi}).

\newpage

\section{The Winning Approach}\label{sec_winning}

\subsection{Data Distribution}

Our task is to predict the destination of a taxi given a prefix of its trajectory. As the dataset is composed of full trajectories, we have to generate trajectory prefixes by cutting the trajectories in the right way. The provided training dataset is composed of more than 1.7 million complete trajectories, which gives 83 480 696 possible prefixes. The distribution of the training prefixes should be as close as possible as that of the provided testing dataset on which we were eventually evaluated. This test set was selected by taking five snapshots of the taxi network activity at various dates and times. This means that the probability that a trajectory appears in the test set is proportional to its length and that, for each entire testing trajectory, all its possible prefixes had an equal probability of being selected in the test set. Therefore, generating a training set with all the possible prefixes of all the complete trajectories of the original training set provides us with a training set which has the same distribution over prefixes (and whole trajectories) as the test set.

\subsection{MLP Architecture}

A Multi-Layer Perceptron (MLP) is a neural net in which each neuron of a given layer is connected to all the neurons of the next layer, without any cycle. It takes as input fixed-size vectors and processes them through one or several hidden layers that compute higher level representations of the input. Finally the output layer returns the prediction for the corresponding inputs. In our case, the input layer receives a representation of the taxi's prefix with associated metadata and the output layer predicts the destination of the taxi (latitude and longitude). We used standard hidden layers consisting of a matrix multiplication followed by a bias and a nonlinearity. The nonlinearity we chose to use is the Rectifier Linear Unit (ReLU)~\cite{glorot2011deep}, which simply computes $\max(0, x)$. Compared to traditional sigmoid-shaped activation functions, the ReLU limits the gradient vanishing problem as its derivative is always one when $x$ is positive. For our winning approach, we used a single hidden layer of 500 ReLU neurons.

\subsection{Input Layer}

One of the first problems we encountered was that the trajectory prefixes are varying-length sequences of GPS points, which is incompatible with the fixed-size input of the MLP. To circumvent (temporarily, see Section~\ref{rnn}) this limitation, we chose to consider only the first $k$ points and last $k$ points of the trajectory prefix, which gives us a total of $2k$ points, or $4k$ numerical values for our input vector. For the winning model we took $k=5$. These GPS points are standardized (zero-mean, unit-variance). When the prefix of the trajectory contains less than $2k$ points, the first and last $k$ points overlap. When the trajectory prefix contains less than $k$ points, we pad the input GPS points by repeating either the first or the last point.

To deal with the discrete meta-data, consisting of client ID, taxi ID, date and time information, we learn embeddings jointly with the model for each of these information. This is inspired by neural language modeling approaches~\cite{bengio2003neural}, in which each word is mapped to a vector space of fixed size (the vector is called the \emph{embedding} of the word). The table of embeddings for the words is included in the model parameters that we learn and behaves as a regular parameter matrix: the embeddings are first randomly initialized and are then modified by the training algorithm like the other parameters. In our case, instead of words, we have metadata values. More precisely, we have one embedding table for each metadata with one row for each possible value of the metadata. For the date and time, we decided to create higher-level variables that better describe human activity: quarters of hour, day of the week, week of the year (one embedding table is learnt for each of them). These embeddings are then simply concatenated to the $4k$ GPS positions to form the input vector of the MLP. The complete list of embeddings used in the winning model is given in Table~\ref{table_embed}.

\begin{table}
\centering
\caption{Metadata values and associated embedding size.}
\label{table_embed}
\begin{tabular}{l||c|c}
  Metadata & Number of possible values & Embedding size \\
  \hline \hline
  Client ID & 57106 & 10\\
  Taxi ID & 448 & 10\\
  Stand ID & 64 & 10\\
  Quarter hour of the day & 96 & 10\\
  Day of the week & 7 & 10\\
  Week of the year & 52 & 10
\end{tabular}
\end{table}

\subsection{Destination Clustering and Output Layer}

As the destination we aim to predict is composed of two scalar values (latitude and longitude), it is natural to have two output neurons. However, we found that it was difficult to train such a simple model because it does not take into account any prior information on the distribution of the data. To tackle this issue, we integrate prior knowledge of the destinations directly in the architecture of our model: instead of predicting directly the destination position, we use a predefined set $(c_i)_{1\le i \le C}$ of a few thousand destination cluster centers and a hidden layer that associates a scalar value $(p_i)_i$ (similar to a probability) to each of these clusters. As the network must output a single destination position, for our output prediction $\hat{y}$, we compute a weighted average of the predefined destination cluster centers:
\[ \hat{y} = \sum_{i=1}^C p_i c_i. \]
Note that this operation is equivalent to a simple linear output layer whose weight matrix would be initialized as our cluster centers and kept fixed during training.
The hidden values $(p_i)_i$ must sum to one so that $\hat{y}$ corresponds to a centroid calculation and thus we compute them using a softmax layer:
\[ p_i = \frac{\exp(e_i)}{\sum_{j=1}^C \exp(e_j)}, \]
where $(e_j)_j$ are the activations of the previous layer.

\begin{figure}
	\centering
	\scalebox{.9}{%
		\def\svgwidth{.8\textwidth}
		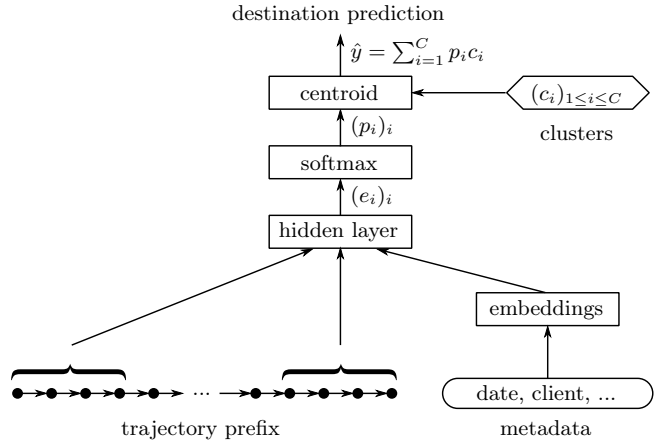
	}
	\caption{Architecture of the winning model.}
	\label{mlp_fig}
\end{figure}

The clusters $(c_i)_i$ were calculated with a mean-shift clustering algorithm on the destinations of all the training trajectories, returning a set of $C=3392$ clusters. Our final MLP architecture is represented in Figure~\ref{mlp_fig}.

\subsection{Cost Computation and Training Algorithm}

The evaluation cost of the competition is the mean Haversine distance, which is defined as follows ($\lambda_x$ is the longitude of point $x$, $\phi_x$ is its latitude, and $R$ is the radius of the Earth):
\[ d_\mathrm{haversine}(x, y) = 2 R \arctan\left(\sqrt{\frac{a(x, y)}{a(x, y)-1}}\right), \]
where $a(x, y)$ is defined as:
\[ a(x, y) = \sin^2\left(\frac{\phi_y-\phi_x}{2}\right)+\cos(\phi_x)\cos(\phi_y)\sin^2\left(\frac{\lambda_y-\lambda_x}{2}\right). \]
Our models did not learn very well when trained directly on the Haversine distance function and thus, we used the simpler  equirectangular distance instead, which is a very good approximation at the scale of the city of Porto:
\[ d_\mathrm{equirectangular}(x, y) = R \sqrt{\left((\lambda_y-\lambda_x)\cos\left(\frac{\phi_y-\phi_x}{2}\right)\right)^2+(\phi_y-\phi_x)^2}. \]

We used stochastic gradient descent (SGD) with momentum to minimise the mean equirectangular distance between our predictions and the actual destination points. We set a fixed learning rate of 0.01, a momentum of 0.9 and a batch size of 200.

\section{Alternative Approaches}\label{sec_alternatives}

The models that we are going to present in this section did not perform as well for our specific destination task on the competition test set but we believe that they can provide interesting insights for other problems involving fixed-length outputs and variable-length inputs.

\subsection{Recurrent Neural Networks}\label{rnn}

\begin{figure}[!ht]
	\centering
	\scalebox{.9}{%
		\def\svgwidth{.8\textwidth}
		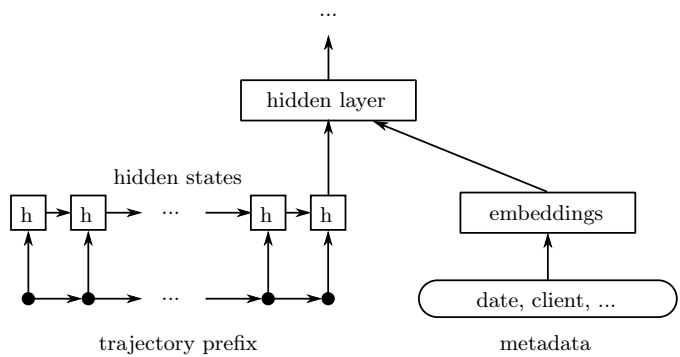
	}
	\caption{RNN architecture, in which the GPS points are read one by one in the forward direction of the prefix. (The rest of the architecture is the same as before and is omitted for clarity.)}
	\label{fig_rnn}
\end{figure}

As stated previously, a MLP is constrained by its fixed-length input, which prevents us from fully exploiting the entire trajectory prefix. Therefore we naturally considered recurrent neural net (RNN) architectures, which can read all the GPS points one by one, updating a fixed-length internal state with the same transition matrix at each time step. The last internal state of the RNN is expected to summarize the prefix with relevant features for the specific task. Such recurrent architectures are difficult to train due in particular to the problem of vanishing and exploding gradients~\cite{bengio1994learning}. This problem is partially solved with long short-term memory (LSTM) units~\cite{hochreiter1997long}, which are crucial components in many state of the art architectures for tasks including handwriting recognition~\cite{graves2009novel, doetsch2014fast}, speech recognition~\cite{graves2013speech, sak2014long}, image captioning~\cite{xu2015show} or machine translation~\cite{bahdanau2014neural}.

We implemented and trained a LSTM RNN that reads the trajectory one GPS point at a time from the beginning to the end of each input prefix. This architecture is represented in Figure~\ref{fig_rnn}. Furthermore, in order to help the network to better identify short-term dependencies (such as the velocity of the taxi as the difference between two successive data points), we also considered a variant in which the input of the RNN is not anymore a single GPS point but a window of 5 successive GPS points of the prefix. The window shifts along the prefix by one point at each RNN time step.

\subsection{Bidirectional Recurrent Neural Networks}

\begin{figure}
	\centering
	\scalebox{.9}{%
		\def\svgwidth{.8\textwidth}
		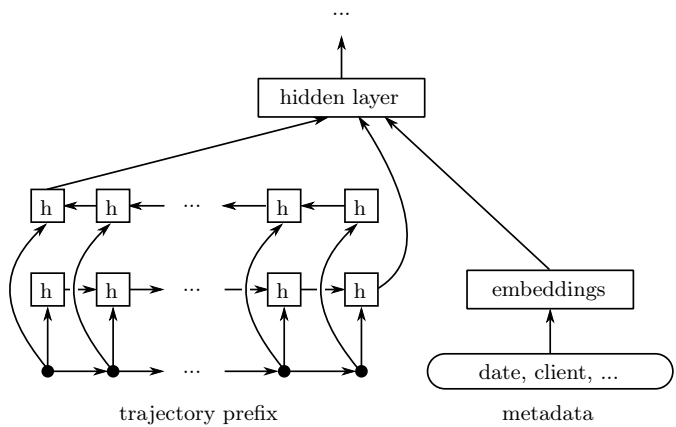
	}
	\caption{Bidirectional RNN architecture.}
	\label{fig_bidirnn}
\end{figure}

We noticed that the most relevant parts of the prefix are its beginning and its end and we therefore tried a bidirectional RNN~\cite{graves2005bidirectional} (BRNN) to focus on these two particular parts. Our previously described RNN reads the prefix forwards from the first to the last known point, which leads to a final internal state containing more information about the last points (the information about the first points is more easily forgotten). In the BRNN architecture, one RNN reads the prefix forwards while a second RNN reads the prefix backwards. The two final internal states of the two RNNs are then concatenated and fed to a standard MLP that will predict the destination in the same way as in our previous models. The concatenation of these two final states is likely to capture more information about the beginning and the end of the prefix. Figure~\ref{fig_bidirnn} represents the BRNN component of our architecture.

\subsection{Memory Networks}

\begin{figure}
	\centering
	\scalebox{.9}{%
		\def\svgwidth{.8\textwidth}
		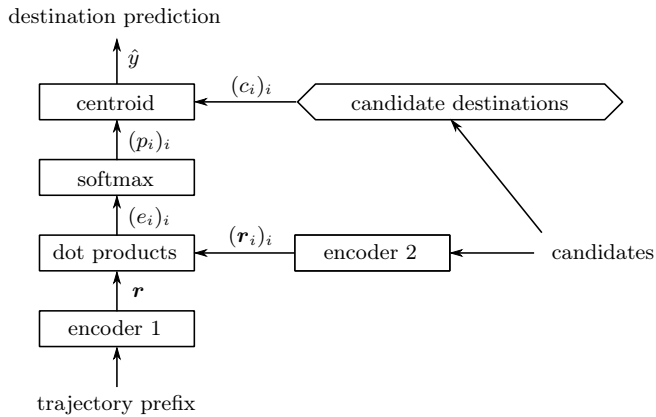
	}
	\caption{Memory network architecture. The \emph{encoders} are generic bricks that take as input the trajectory points with metadata and process them through a feedforward or recurrent neural net in order to return a fixed-size representations $\vec{r}$ and $(\vec{r})_i$.}
	\label{fig_memnet}
\end{figure}

Memory networks~\cite{weston2014memory} have been recently introduced as an architecture that can exploit an external database by retrieving and storing relevant information for each prediction. We have implemented a slightly related architecture, which is represented in Figure~\ref{fig_memnet} and which we will now describe. For each prefix to predict, we extract $m$ entire trajectories (which we call \emph{candidates}) from the training dataset. We then use two neural network encoders to respectively encode the prefix and the candidates (same encoder for all the candidates). The encoders are the same as those of our previous architectures (either feedforward or recurrent) except that we stop at the hidden layer instead of predicting an output. This results into $m+1$ fixed-length representations in the same vector space so that they can be easily compared. Then we compute similarities by taking the dot products of the prefix representation with all the candidate representations. Finally we normalize these $m$ similarity values with a softmax and use the resulting probabilities to weigh the destinations of the corresponding candidates. In other words, the final destination prediction of the prefix is the centroid of the candidate destinations weighted by the softmax probabilities. This is similar to the way we combine clusters in our previously described architectures.

\vspace{-2em}

As the trajectory database is very large, such an architecture is quite challenging to implement efficiently. Therefore, for each prefix (more precisely for each batch of prefixes), we naively select $m=10000$ random candidates. We believe that more sophisticated retrieving functions could significantly improve the results, but we did not have time to implement them. In particular, one could use a pre-defined (hand-engineered) similarity measure to retrieve the most similar candidates to the particular prefix.

The two encoders that map prefixes and candidates into the same representation space can either be feedforward or recurrent (bidirectional). As RNNs are more expensive to train (both in terms of computation time and RAM consumption), we had to limit ourselves to a MLP with one single hidden layer of 500 ReLUs for the encoder. We trained the architecture with a batch size of 5000 examples, and for each batch we randomly pick 10000 candidates from the training set.

\section{Experimental Results}\label{sec_exp}

\subsection{Custom Validation Set}

As the competition testing dataset is particularly small, we can not reliably compare models on it. Therefore, for the purpose of this paper, we will compare our models on two bigger datasets: a validation dataset composed of 19427 trajectories and a testing dataset composed of 19770 trajectories. We obtained these new testing and validation sets by extracting (and removing) random portions of the original training set. The validation dataset is used to early-stop our training algorithms for each model based on the best validation score, while the testing dataset is used to compare our different trained models.

\begin{table}[bh]
\centering
\caption{Testing errors of our models. Contrary to model 1, model 2 predicts the destination directly without using the clusters. Model 3 only uses the prefix as input without any metadata. Model 4 only uses metadata as input without the prefix. While the BRNN of model 6 takes a single GPS point at each time step, the BRNN of model 7 takes five consecutive GPS points at each time step.}
\label{table_results}
\begin{tabular}{l l||c|c|c}
  & Model & Custom Test & Kaggle Public & Kaggle Private \\
  \hline \hline
  1 & MLP, clustering (winning model) & 2.81 & \textbf{2.39} & \textbf{1.87}\footnotemark[8] \\
  2 & MLP, direct output & 2.97 & 3.44 & 3.88 \\
  3 & MLP, clustering, no embeddings & 2.93 & 2.64 & 2.17 \\
  4 & MLP, clustering, embeddings only & 4.29 & 3.52 & 3.76 \\
  5 & RNN & 3.14 & 2.49 & 2.39 \\
  6 & Bidirectional RNN & 3.01 & 2.65 & 2.33 \\
  7 & Bidirectional RNN with window & \textbf{2.60} & 3.15 & 2.06 \\
  8 & Memory network & 2.87& 2.77 & 2.20 \\
   & Second-place team & - & 2.36 & 2.09 \\
   & Third-place team & - & 2.45 & 2.11 \\
   & Average competition scores\footnotemark[9] & - & 3.02 & 3.11 \\
\end{tabular}
\end{table}

\footnotetext[8]{our winning submission on Kaggle scored 2.03 but the model had not been trained until convergence}
\footnotetext[9]{average over 381 teams, the submissions with worse scores than the public benchmark (in which the center of Porto is always predicted) have been discarded}

\subsection{Results}

Table~\ref{table_results} shows the testing scores of our various models on our custom testing dataset as well as on the competition ones. The different hyperparameters of each model have been tuned.

\section{Analysis of the results}\label{sec_dis}

Our winning ECML/PKDD challenge model is the MLP model that uses clusters of the destinations. However it is not our best model on our custom test, which is the BRNN with window. As our custom test set is considerably larger than the competition one, the scores on it are significantly more confident and we can therefore assert that our overall best model is the BRNN with window.

\begin{figure}[!b]
\centering
\begin{subfigure}{.5\textwidth}
  \centering
  \includegraphics[width=\linewidth]{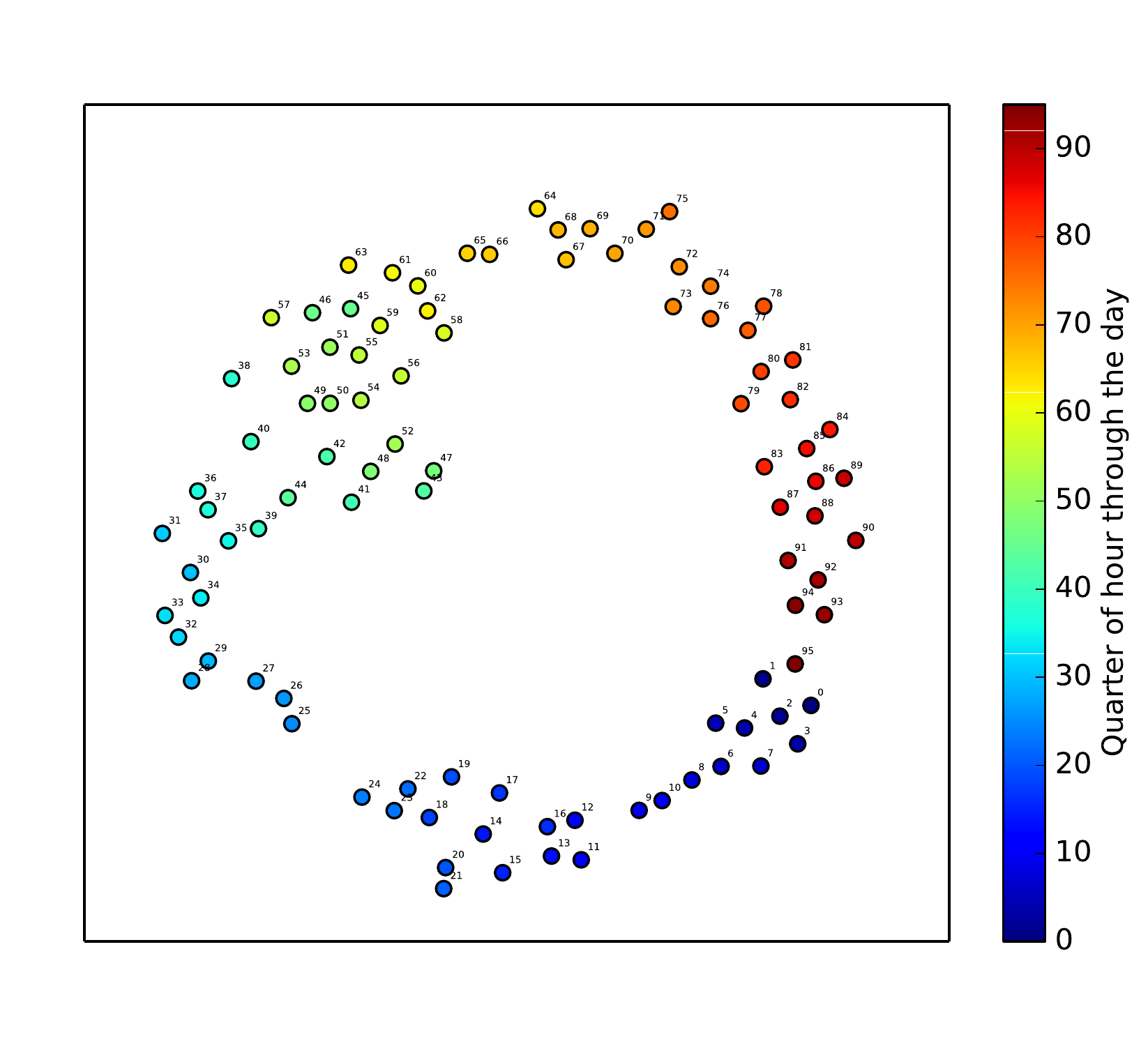}
  \label{fig_qhour}
\end{subfigure}%
\begin{subfigure}{.5\textwidth}
  \centering
  \includegraphics[width=\linewidth]{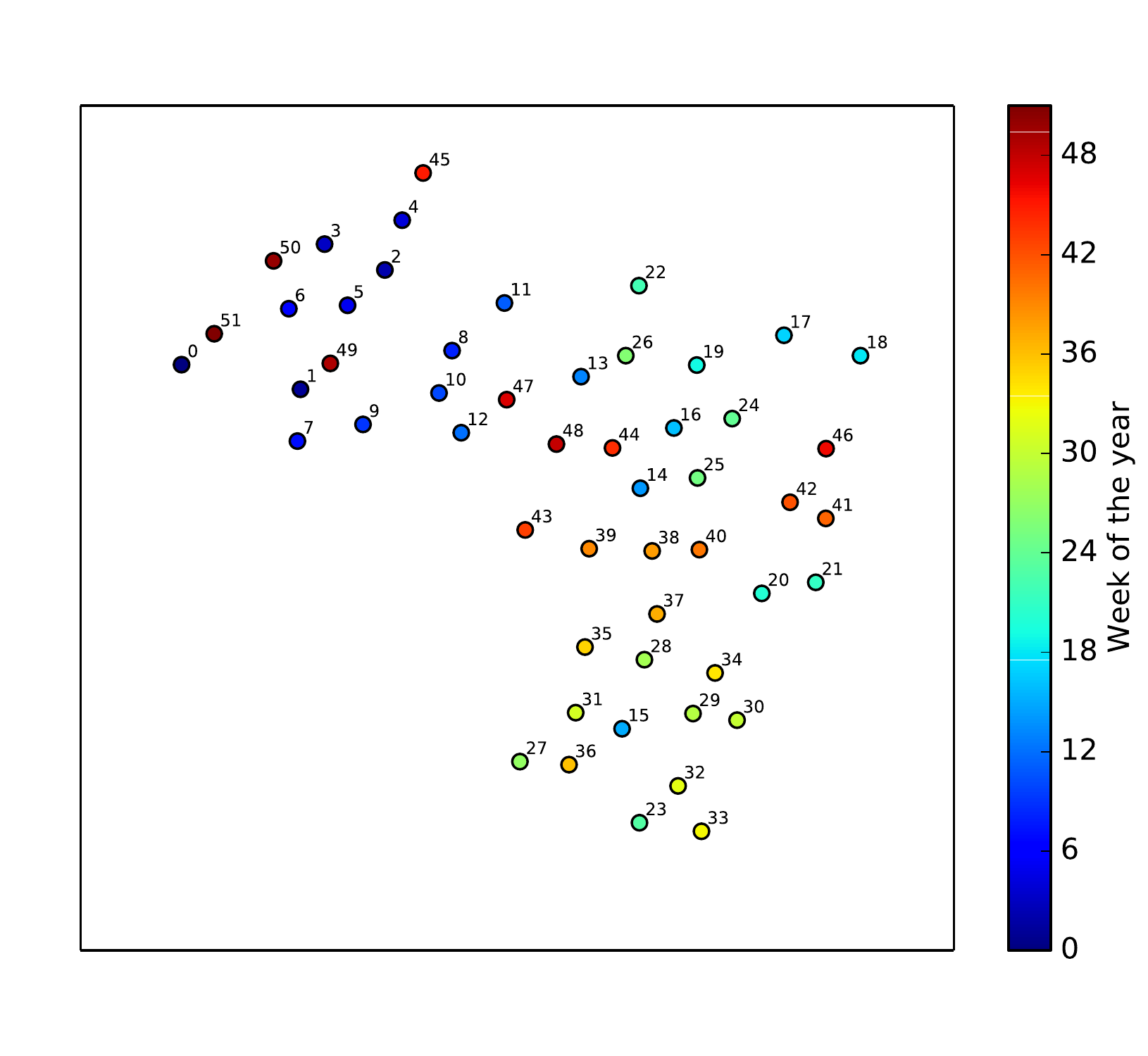}
  \label{fig_week}
\end{subfigure}
\vspace{-2em}
\caption{t-SNE 2D projection of the embeddings. \textbf{Left:} quarter of hour at which the taxi departed (there are 96 quarters in a day, so 96 points, each one representing a particular quarter). \textbf{Right:} week of the year during which the taxi departed (there are 52 weeks in a year, so 52 points, each one representing a particular week).}
\label{fig_tsne}
\end{figure}

The results also prove that embeddings and clusters significantly improve our models. The importance of embeddings can also be confirmed by visualizing them. Figure~\ref{fig_tsne} shows 2D t-SNE~\cite{van2008visualizing} projections for two of these embeddings and clear patterns can be observed, proving that quarters of hour and weeks of the year are important features for the prediction.

The reported score of the memory network is lower than the others but this might be due to the fact that it was not trained until convergence (we stopped after one week on a high end GPU).

The scores on our custom test set are higher than the scores on the public and private test set used for the competition. This suggests that the competition testing set is composed of rides that took place at very specific dates and times with very particular trajectory distributions. The gap between the public and private test sets is probably due to the fact that their size is particularly small. In contrast, our validation and test sets are big enough to obtain more significant statistics.

All the models we have explored are very computationally intensive and we thus had to train them on GPUs to avoid weeks of training. Our competition winning model is the least intensive and can be trained in half a day on GPU. On the other hand, our recurrent and memory networks are much slower and we believe that we could reach even better scores by training them longer.

\section*{Conclusion}

We introduced an almost fully-automated neural network approach to predict the destination of a taxi based on the beginning of its trajectory and associated metadata. Our best model uses a recurrent bidirectional neural network to encode the prefix, several embeddings to encode the metadata and destination clusters to generate the output.

One potential limitation of our clustering-based output layer is that the final prediction can only fall in the convex hull of the clusters. A potential solution would be to learn the clusters as parameters of the network and initialize them either randomly or from the mean-shift clusters.

Concerning the memory network, one could consider more sophisticated ways to extract candidates, such as using an hand-engineered similarity measure or even the similarity measure learnt by the memory network. In this latter case, the learnt similarity should be used to extract only a proportion of the candidates in order let a chance to candidates with poor similarities to be selected. Furthermore, instead of using the dot product to compare prefix and candidate representations, more complex functions could be used (such as the concatenation of the representations followed by non-linear layers).

\subsection*{Acknowledgments}
We would like to thank the developers of Theano \cite{theano0, theano1}, Blocks and Fuel \cite{blocksfuel} for developing such convenient tools.
We acknowledge the support of the following organizations for research funding and computing support: Samsung, NSERC, Calcul Quebec, Compute Canada, the Canada Research Chairs and CIFAR.

\bibliography{taxi}
\bibliographystyle{unsrt}

\end{document}